\documentclass{article}
\usepackage{ijcai23}

\usepackage{times}
\usepackage{soul}
\usepackage{url}
\usepackage[hidelinks]{hyperref}
\usepackage[utf8]{inputenc}
\usepackage[small]{caption}
\usepackage{graphicx}
\usepackage{amsmath}
\usepackage{amsthm}
\usepackage{booktabs}
\usepackage{algorithm}
\usepackage{algorithmic}
\usepackage[switch]{lineno}

\newtheorem{theorem}{Theorem}

\title{Causal Effect Estimation with Variational AutoEncoder \\and the Front Door Criterion}

\author{
Ziqi Xu$^1$
\and
Debo Cheng$^1$\and
Jiuyong Li$^1$\and
Jixue Liu$^1$\and
Lin Liu$^1$\And
Kui Yu$^2$
\affiliations
$^1$University of South Australia\\
$^2$Hefei University of Technology
\emails
\{ziqi.xu, debo.cheng\}@mymail.unisa.edu.au,\{jiuyong.li,jixue.liu,lin.liu\}@unisa.edu.au,
yukui@hfut.edu.cn}

\usepackage{array}
\usepackage{amsfonts}       
\usepackage{amsthm}
\usepackage{amsmath}
\usepackage{algorithm}
\usepackage{algorithmic}
\usepackage{theoremref}
\usepackage{xcolor}
\usepackage{titlesec}
\usepackage{multirow}
\usepackage{diagbox}
\usepackage{threeparttable}
\usepackage{verbatim}
\usepackage{booktabs}
\usepackage{subcaption}
\usepackage{setspace}

\newtheorem{definition}{Definition}
\newtheorem{assumption}{Assumption}

\newtheorem{problem}{Problem}

\makeatletter
\newcommand*{\indep}{%
	\mathbin{%
		\mathpalette{\@indep}{}%
	}%
}
\newcommand*{\nindep}{%
	\mathbin{% The final symbol is a binary math operator
		\mathpalette{\@indep}{\not}% \mathpalette helps for the adaptation
	}%
}
\newcommand*{\@indep}[2]{%
	\sbox0{$#1\perp\m@th$}%box 0 contains \perp symbol
	\sbox2{$#1=$}%box 2 for the height of =
	\sbox4{$#1\vcenter{}$}% box 4 for the height of the math axis
	\rlap{\copy0}% first \perp
	\dimen@=\dimexpr\ht2-\ht4-.2pt\relax
	\kern\dimen@
	{#2}%
	\kern\dimen@
	\copy0 %
}

\begin{document}

\maketitle

\begin{abstract}
	An essential problem in causal inference is estimating causal effects from observational data. The problem becomes more challenging with the presence of unobserved confounders. When there are unobserved confounders, the commonly used back-door adjustment is not applicable. Although the instrumental variable (IV) methods can deal with unobserved confounders, they all assume that the treatment directly affects the outcome, and there is no mediator between the treatment and the outcome. This paper aims to use the front-door criterion to address the challenging problem with the presence of unobserved confounders and mediators. In practice, it is often difficult to identify the set of variables used for front-door adjustment from data. By leveraging the ability of deep generative models in representation learning, we propose FDVAE to learn the representation of a \underline{F}ront-\underline{D}oor adjustment set with a \underline{V}ariational \underline{A}uto\underline{E}ncoder, instead of trying to search for a set of variables for front-door adjustment. Extensive experiments on synthetic datasets validate the effectiveness of FDVAE and its superiority over existing methods. The experiments also show that the performance of FDVAE is not sensitive to the causal strength of unobserved confounders and is feasible in the case of dimensionality mismatch between learned representations and the ground truth. We further apply the method to three real-world datasets to demonstrate its potential applications.
\end{abstract}

\section{Introduction}
Estimating causal effects is a fundamental problem in many application areas. Firms need to estimate the effects of different marketing strategies and identify the best ones to achieve high profits~\cite{ascarza2018retention}. Policymakers need to know whether the implementation of a policy has a positive impact on the community~\cite{athey2017beyond,tran2022most}. Medical researchers study the effects of treatments on patients~\cite{petersen2014causal}. Causal inference is also crucial to understanding the nature of how things develop~\cite{bareinboim2016causal,peters2017elements}.

Randomised Controlled Trials (RCTs)~\cite{fisher1936design} are considered the golden rule for estimating causal effects. However, RCTs are difficult to implement in many real-world cases due to ethical issues or huge costs~\cite{deaton2018understanding}. For example, it would be unethical to subject an individual to a condition if the condition may have potentially negative consequences (e.g., smoking). Therefore, many methods have been developed to infer causal effects from observational data. Most of the methods assume no unobserved confounders (variables affecting both the treatment and outcome), i.e., make the unconfoundedness assumption~\cite{imbens2015causal}, and follow the back-door criterion~\cite{pearl2009causality} to determine valid adjustment sets to control confounders for unbiased estimation.

However, unobserved confounders are commonplace in practice. Currently, only a few instrumental variables (IV) based methods are available for handling unobserved confounders, but an IV needs to satisfy the following conditions~\cite{hernan2006instruments} (which restrict the applications of IV based methods): (i) correlating with the treatment (i.e., relevance condition); (ii) affecting the outcome only through the treatment (i.e., exclusion restriction); and (iii) no confounding bias\footnote{Here no confounding bias means there do not exist variables that causally affect both the IV and the outcome.} between the IV and the outcome (i.e., unconfounded instrument). We note that the second assumption is too restrictive in practice, since quite often there may exist a mediator between the treatment and the outcome. For instance, lung cancer development is not directly affected by smoking cigarettes but is mediated through tar deposits.

\begin{figure}[t]
	\centering
	\includegraphics[scale=0.45]{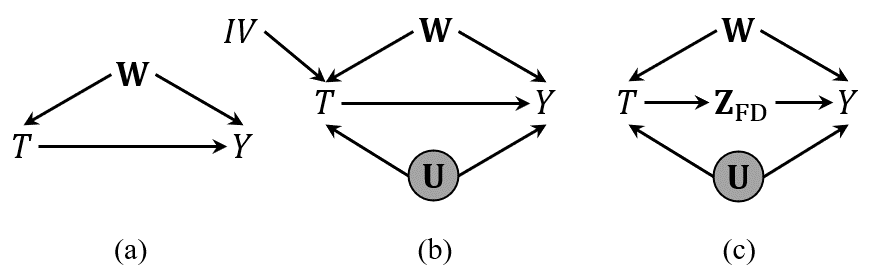}
	\caption{Causal graphs of some models. $T$ is a treatment; $Y$ is an outcome; $\mathbf{W}$ is a set of confounders; $\mathbf{U}$ is a set of unobserved confounders; $IV$ is a instrumental variable; and $\mathbf{Z}_\mathrm{FD}$ is a set of variables that satisfy the front-door criterion.}
	\label{pic:intro}
\end{figure}

The key to causal effect estimation from observational data is to determine the identifiability of a causal effect based on the data. That is, the causal effect identification problem asks whether the effect of holding the treatment $T$ at a constant value $t$ on the outcome $Y$, written as $P(Y|do(T=t))$, or $P(Y|do(t))$, can be computed from a combination of observational data and causal assumptions. Let us use examples to show the different cases. The causal DAG (directed acyclic graph) shown in Figure~\ref{pic:intro}(a) is a simple case that satisfies the unconfoundedness assumption. The value of $P(Y|do(t))$ can be estimated by the back-door adjustment formula~\cite{pearl2009causality}. However, when there exist unobserved confounders, only adjusting on $\mathbf{W}$ does not block all back-door paths from $T$ to $Y$. Hence, back-door criterion based methods cannot handle the cases in Figure~\ref{pic:intro}(b) or \ref{pic:intro}(c). With the case in Figure~\ref{pic:intro}(b) though, the causal effect of $T$ on $Y$ can be unbiasedly estimated when an IV is given. However, an IV based method cannot solve the case in Figure~\ref{pic:intro}(c) with mediators. In this paper, we aim to solve the problem in Figure~\ref{pic:intro}(c), i.e., the exclusion restriction does not hold and there exist unobserved confounders. 

With a case like that in Figure~\ref{pic:intro}(c), the front-door adjustment can be applied to obtain unbiased causal effect estimation. However, in practice, it is challenging to identify from data a suitable adjustment set satisfying the front-door criterion. Only a few researchers have paid attention to this problem. For example, Jeong et al.~\shortcite{jeong2022finding} proposed a search algorithm that can find a suitable front-door adjustment set, but with the assumption that there must exist such an adjustment set in a given causal graph. Wien{\"o}bst et al.~\shortcite{wienobst2022finding} improved the above algorithm to reduce the time complexity from polynomial time to linear time. The above algorithms still have some drawbacks, because they are not data-driven and need to give a causal graph, so they tend to be inefficient when dealing with high-dimensional data.

Recently, deep generative models based on Variational AutoEncoder (VAE)~\cite{KingmaW13,ZhangLL21,cheng2022causal} have achieved many successes in representation learning for causal inference~\cite{LouizosSMSZW17}. For instance, Louizos et al.~\shortcite{LouizosSMSZW17} first combined causal inference and VAE, and proposed CEVAE. CEVAE can learn the representation of latent confounders as a back-door adjustment set for estimating the average treatment effect (ATE). Cheng et al.~\shortcite{cheng2022causal} proposed a VAE based method to estimate the natural direct effect and the natural indirect effect for mediation analysis.

As it is challenging to identify a suitable front-door adjustment set directly, we propose to learn the representation of a front-door adjustment set from the proxy variables by using the VAE technique. For a treatment $T$ and an outcome $Y$, a set of unobserved confounders $\mathbf{U}$, and a set of proxy variables $\mathbf{X}$ which is the proxy of $\mathbf{Z}_\mathrm{FD}$ and $\mathbf{W}$, where $\mathbf{Z}_\mathrm{FD}$ is a set of variables that satisfy the front-door criterion and $\mathbf{W}$ is a set of confounders, we propose the representation learning strategy under the causal graph as shown in Figure~\ref{pic:Problem} to learn $\boldsymbol{\Psi}$, the representation of $\mathbf{Z}_\mathrm{FD}$ from $\mathbf{X}$. We develop a VAE based method, namely FDVAE (Front-Door adjustment set learning based on Variational AutoEncoder) to learn $\boldsymbol{\Psi}$ for unbiased average treatment effect estimation. This method can unbiasedly estimate causal effect in the presence of unobserved confounders and mediators, and it is the first data-driven method based on the front-door criterion. The contributions of this paper can be summarised as follows:

\begin{itemize}
	\item We study a practical case of causal effect estimation when there exist mediators between the treatment and the outcome and there are unobserved confounders.
	
	\item We propose a novel causal effect estimation method, FDVAE, to learn the representation of a front-door adjustment set from proxy variables to more accurately estimate the average treatment effect.
	
	\item We evaluate the effectiveness of the FDVAE method on synthetic datasets. Experiments show that FDVAE outperforms existing methods. Furthermore, we apply FDVAE to three real-world datasets to show the application scenarios of FDVAE.
\end{itemize}

The rest of this paper is organised as follows. In Section 2, we discuss the preliminaries for causal inference. The details of FDVAE are presented in Section 3. In Section 4, we discuss the experiment results. In Section 5, we discuss related works. Finally, we conclude the paper in Section 6.

\section{Preliminaries}
In this section, we present the necessary background of causal inference. We use a capital letter to represent a variable and a lowercase letter to represent its value. Boldfaced capital and lowercase letters are used to represent sets of variables and values, respectively. 

Let $\mathcal{G}=(\mathbf{V}, \mathbf{E})$ be a directed acyclic graph (DAG), where $\mathbf{V}=\{V_{1}, \dots, V_{p}\}$ is the set of nodes and $\mathbf{E}$ is the set of edges between the nodes, i.e., $\mathbf{E}\subseteq \mathbf{V}\times \mathbf{V}$. A path $\pi$ from $V_{s}$ to $V_{e}$ is a sequence of distinct nodes $<V_{s}, \dots, V_{e}>$ such that every pair of successive nodes are adjacent in $\mathcal{G}$. In $\mathcal{G}$, if there exists $V_i\rightarrow V_j$, $V_i$ is a parent of $V_j$ and we use $Pa(V_j)$ to denote the set of all parents of $V_j$.

We follow Pearl's work~\cite{pearl2009causality} and use Structural Causal Models (SCMs) as our basic framework. A SCM is a triple $(\mathbf{U},\mathbf{V},\mathbf{F})$, where $\mathbf{U}$ is a set of exogenous (unobserved) variables; $\mathbf{V}$ is a set of endogenous (observed) variables; and $\mathbf{F}$ is a set of deterministic functions, {\small $V_i = f_i(Pa(V_i),U_{Pa(V_i)})$}, such that {\small $Pa(V_i)\subseteq \mathbf{V}\backslash\{V_i\}$} and {\small $U_{Pa(V_i)}\subseteq \mathbf{U}$}.

\begin{assumption} [Markov Condition~\cite{pearl2009causality}]
	\label{Markovcondition}
	Given a DAG $\mathcal{G}=(\mathbf{V}, \mathbf{E})$ and $P(\mathbf{V})$, the joint probability distribution of $\mathbf{V}$, $\mathcal{G}$ satisfies the Markov Condition if for $\forall V_i \in \mathbf{V}$, $V_i$ is probabilistically independent of all of its non-descendants, given the parent nodes of $V_i$.
\end{assumption}

With the Markov Condition, $P(\mathbf{V})$ can be factorised by: $P(\mathbf{V}) = \prod_{i} P(V_i|Pa(V_i))$.

\begin{assumption}[Faithfulness~\cite{spirtes2000causation}]
	\label{Faithfulness}
	A DAG {\small $\mathcal{G}=(\mathbf{V}, \mathbf{E})$} is faithful to {\small $P(\mathbf{V})$} iff every independence present in {\small $P(\mathbf{V})$} is entailed by {\small $\mathcal{G}$} and satisfies the Markov Condition. {\small $P(\mathbf{V})$} is faithful to {\small $\mathcal{G}$} iff there exists {\small $\mathcal{G}$} which is faithful to {\small $P(\mathbf{V})$}.
\end{assumption}

When the Markov Condition and Faithfulness assumption are satisfied, we can use $d$-separation defined below to read the conditional independence between variables entailed in the DAG $\mathcal{G}$.

\begin{definition}[$d$-separation~\cite{pearl2009causality}]
	\label{d-separation}
	A path $\pi$ in a DAG is said to be $d$-separated (or blocked) by a set of nodes $\mathbf{Z}$ iff (1) the path $\pi$ contains a chain $V_i \rightarrow V_k \rightarrow V_j$ or a fork $V_i \leftarrow V_k \rightarrow V_j$ such that the middle node $V_k$ is in $\mathbf{Z}$, or (2) the path $\pi$ contains an inverted fork (or collider) $V_i \rightarrow V_k \leftarrow V_j$ such that $V_k$ is not in $\mathbf{Z}$ and no descendant of $V_k$ is in $\mathbf{Z}$.
\end{definition}

The goal of this paper is to estimate the average treatment effect as defined below.

\begin{definition}[Average Treatment Effect (ATE)]
	\label{def:ATE}
	The average treatment effect of a treatment, denoted as $T$ on the outcome of interest, denoted as $Y$ is defined as $ATE=\mathbb{E}(Y \mid do(T=1))-\mathbb{E}(Y \mid do(T=0))$, where $do()$ is the $do$-operator and $do(T=t)$ represents the manipulation of $T$ by setting its value to $t$~\cite{pearl2009causality}.
\end{definition} 

When the context is clear, we abbreviate $do(T=t)$ as $do(t)$. In order to allow the above $do()$ expressions to be recovered from data, Pearl formally defined causal effect identifiability~\cite{pearl2009causality} (p.77) and proposed two well-known identification conditions, the back-door criterion and front-door criterion.

\begin{definition}[Back-Door (BD) Criterion~\cite{pearl2009causality}]
	\label{def:BD}
	A set of variables $\mathbf{Z}_\mathrm{BD}$ satisfies the back-door criterion relative to an ordered pair of variables $(T,Y)$ in a DAG $\mathcal{G}$ if: (1) no node in $\mathbf{Z}_\mathrm{BD}$ is a descendant of $T$; and (2) $\mathbf{Z}_\mathrm{BD}$ blocks every path between $T$ and $Y$ that contains an arrow into $T$.
\end{definition}

A back-door path is a non-causal path from $T$ to $Y$ which would remain if remove any arrows pointing out of $T$ (these are the potentially causal paths from $T$. They have been recognised as ``back-door” paths because they flow backwards out of $T$, i.e., all of these paths point into $T$.

\begin{theorem}[Back-Door (BD) Adjustment~\cite{pearl2009causality}]
	If $\mathbf{Z}_\mathrm{BD}$ satisfies the BD criterion relative to $(T,Y)$, then the causal effect of $T$ on $Y$ is identifiable and is given by the following BD adjustment formula~\cite{pearl2009causality}:
	\begin{equation}
		\begin{aligned}
			P(y|do(t)) = \sum_{\mathbf{z}_{\mathrm{BD}}}^{}P(y|t,\mathbf{z}_{\mathrm{BD}})P(\mathbf{z}_{\mathrm{BD}}).
		\end{aligned}	
	\end{equation}
\end{theorem}

\begin{definition}[Front-Door (FD) Criterion~\cite{pearl2009causality}]
	\label{def:FD}
	A set of variables $\mathbf{Z}_\mathrm{FD}$ is said to satisfy the front-door criterion relative to an ordered pair of variables $(T,Y)$ in a DAG $\mathcal{G}$ if: (1) $\mathbf{Z}_\mathrm{FD}$ intercepts all directed paths from $T$ to $Y$; (2) there is no unblocked back-door path from $T$ to $\mathbf{Z}_\mathrm{FD}$; and (3) all back-door paths from $\mathbf{Z}_\mathrm{FD}$ to Y are blocked by $T$.
\end{definition}

\begin{theorem}[Front-Door (FD) Adjustment~\cite{pearl2009causality}]
	If $\mathbf{Z}_\mathrm{FD}$ satisfies the FD criterion relative to $(T,Y)$, then the causal effect of $T$ on $Y$ is identifiable and is given by the following FD adjustment formula~\cite{pearl2009causality}:
	\begin{equation}
		\label{eqa:FD2009}
		\begin{aligned}
			P(y|do(t)) = \sum_{\mathbf{z}_{\mathrm{FD}}}^{}P(\mathbf{z}_{\mathrm{FD}}|t)\sum_{t'}^{}P(y|t',\mathbf{z}_{\mathrm{FD}})P(t'),
		\end{aligned}	
	\end{equation}where $t'$ is a distinct realisation of $T$.
\end{theorem}

\section{The Proposed FDVAE Method}
\begin{figure}[t]
	\centering
	\includegraphics[scale=0.45]{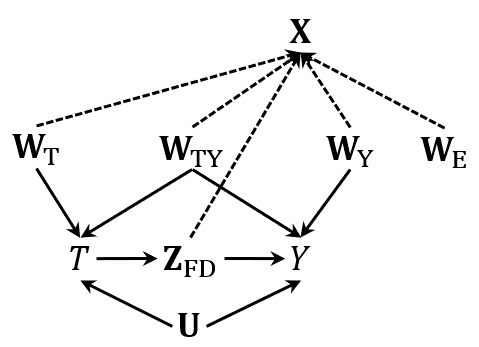}
	\caption{An example causal graph that represents the data generation mechanism.}
	\label{pic:Problem}
\end{figure}

\subsection{Problem Setup}
We assume the data generation is based on the DAG $\mathcal{G} = (\mathbf{U}\cup\mathbf{X}\cup\{T,Y\},\mathbf{E})$ as shown in Figure~\ref{pic:Problem}, which contains the treatment variable $T$, the outcome variable $Y$, the set of unobserved confounders $\mathbf{U}$ and the set of proxy variables $\mathbf{X}$. $\mathbf{X}$ is the proxy of $\mathbf{Z}_\mathrm{FD}$ and $\mathbf{W}$, where $\mathbf{Z}_\mathrm{FD}$ denotes the set that satisfies the FD criterion, i.e., it is a proper FD adjustment set,  {\small $\mathbf{W}=\{\mathbf{W}_{\mathrm{T}},\mathbf{W}_{\mathrm{TY}},\mathbf{W}_{\mathrm{Y}},\mathbf{W}_{\mathrm{E}}\}$}, and {\small $\mathbf{W}_{\mathrm{T}},\mathbf{W}_{\mathrm{TY}},\mathbf{W}_{\mathrm{Y}}$} and {\small $\mathbf{W}_{\mathrm{E}}$} denote four types of variables affecting $T$, affecting both $T$ and $Y$, affecting $Y$, and external variable, respectively. 

Note that as FDVAE aims to learn a representation from the set of proxy variables $\mathbf{X}$ which captures the appropriate information of an adjustment set satisfying the front-door criterion, as long as $\mathbf{X}$ is a set of proxy variables of $\mathbf{Z}_\mathrm{FD}$, FDVAE is able to learn the representation of the adjustment set from $\mathbf{X}$, and whether $\mathbf{Z}_\mathrm{FD}$ and $\mathbf{W}$ are observed or not is irrelevant to the success of the representation learning. Our goal is to query the ATE of $T$ on $Y$ from observational data.

In an ideal scenario, if $\mathbf{Z}_\mathrm{FD}$ can be identified from $\mathbf{X}$, we can use $\mathbf{Z}_\mathrm{FD}$ to estimate the causal effect of $T$ on $Y$ as follows using the FD adjustment formula (here we assume both $T$ and $Y$ are binary):
\begin{equation*}
	{\small\begin{aligned}
		ATE&=\mathbb{E}(Y \mid do(T=1))-\mathbb{E}(Y \mid do(T=0))
		 \\&=\Big( \sum_{\mathbf{z}_{\mathrm{FD}}}^{}P(\mathbf{z}_{\mathrm{FD}}|T=1) - \sum_{\mathbf{z}_{\mathrm{FD}}}^{}P(\mathbf{z}_{\mathrm{FD}}|T=0) \Big)\\ &~~~~~~\sum_{y,t'}^{}yP(y|t',\mathbf{z}_{\mathrm{FD}})P(t'). 
	\end{aligned}	}
\end{equation*}

However, in general, we can only access $\{T,Y,\mathbf{X}\}$. The BD criterion cannot be used since there exists the set of unobserved confounders $\mathbf{U}$ that $d$-connects $T$ and $Y$. It is also challenging to identify $\mathbf{Z}_\mathrm{FD}$ from observational data, since the information of $\mathbf{Z}_\mathrm{FD}$ is embedded in proxy variables $\mathbf{X}$. Fortunately, inspired by VAE technique~\cite{KingmaW13} we propose to learn the representation of $\mathbf{Z}_\mathrm{FD}$, denoted as $\boldsymbol{\Psi}$, through the joint probability distribution $P(\mathbf{X},T,Y)$. Our problem setting is defined as follows:

\begin{figure*}[t]
	\centering
	\begin{subfigure}[b]{0.5788\textwidth}
		\centering
		\includegraphics[scale=0.4]{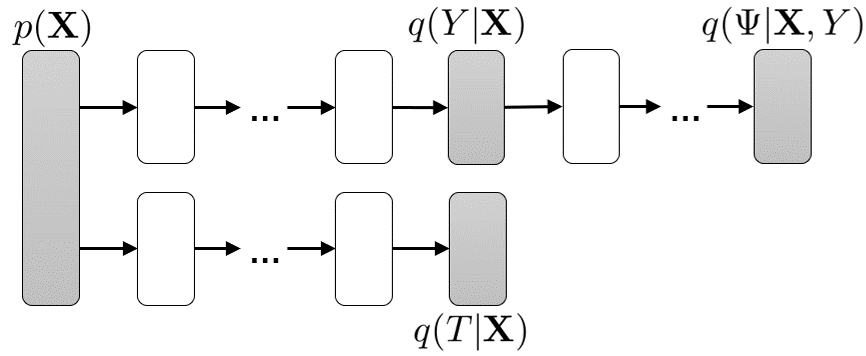}
		\caption{Inference model.}
		\label{pic:FDVAE_A}
	\end{subfigure}
	\begin{subfigure}[b]{0.3712\textwidth}
		\centering
		\includegraphics[scale=0.4]{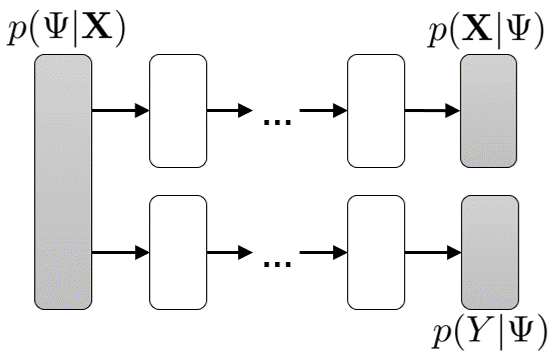}
		\caption{Generative model.}
		\label{pic:FDVAE_B}
	\end{subfigure}
	\caption{The architecture of FDVAE. White nodes represent parameterised deterministic neural network transitions, and grey nodes represent sampling from the respective distributions.}
	\label{pic:FDVAE}
\end{figure*}

\begin{problem}
	Given a joint probability distribution {\small $P(\mathbf{X},T,Y)$} that is generated from the underlying DAG {\small $\mathcal{G} = (\mathbf{U}\cup\mathbf{X}\cup\{T,Y\},\mathbf{E})$} as shown in Figure~\ref{pic:Problem}. Suppose that $\mathbf{X}$ is the proxy of $\mathbf{Z}_\mathrm{FD}$ and $\mathbf{W}$. Our goal is to learn the representation $\boldsymbol{\Psi}$ of $\mathbf{Z}_\mathrm{FD}$ and use $\Psi$ as a FD adjustment set in the estimation of the causal effect of $T$ on $Y$.
\end{problem}

\subsection{Learning Strategy}
In this section, we introduce our method FDVAE for learning the representation $\boldsymbol{\Psi}$ as a proper FD adjustment set from observational data. FDVAE parameterises the causal graph in Figure~\ref{pic:Problem} as a model with neural network functions that connect the variables of interest. We follow the similar learning process in~\cite{KingmaW13}, and the architecture of FDVAE is shown in Figure~\ref{pic:FDVAE}.

In the inference model, we design an encoder $q(\boldsymbol{\Psi} | \mathbf{X},Y)$ that serves as the variational posterior over the target representation. The variational approximation of the posterior is defined as follows:
\begin{equation*}
	{\small\begin{aligned}
		q(\boldsymbol{\Psi} | \mathbf{X},Y) = \prod_{j=1}^{D_{\boldsymbol{\Psi}}}\mathcal{N}(\mu = \hat{\mu}_{\boldsymbol{\Psi}_j}, \sigma^2 = \hat{\sigma}^2_{\boldsymbol{\Psi}_j}),
	\end{aligned}}
\end{equation*}where $D_{\boldsymbol{\Psi}}$ is the dimensionality of $\boldsymbol{\Psi}$, $\hat{\mu}_{\boldsymbol{\Psi}}$ and $\hat{\sigma}^2_{\boldsymbol{\Psi}}$ are the mean and variance of the Gaussian distribution parameterised by neural networks.

Let $D_{\mathbf{X}}$ be the dimensionality of $\mathbf{X}$ and each $g_k(\cdot)$, $k \in \{1,...,8\}$ denotes a neural network parameterised by its own parameters. As shown in Figure~\ref{pic:Problem}, $\mathbf{Z}_\mathrm{FD}$ is determined by $T$, and $T$ is determined by $\{\mathbf{W}_{\mathrm{T}},\mathbf{W}_{\mathrm{TY}}\}$. We choose $\mathbf{X}$ as the condition to regulate the generation process of $\boldsymbol{\Psi}$, since $\mathbf{X}$ is the proxy of $\{\mathbf{W}_{\mathrm{T}},\mathbf{W}_{\mathrm{TY}}\}$. The generative model for $\boldsymbol{\Psi}$ is defined as:
\begin{equation*}
{\small 	\begin{aligned}
		p(\boldsymbol{\Psi} | \mathbf{X}) = \prod_{j=1}^{D_{\boldsymbol{\Psi}}}\mathcal{N}(\mu = \hat{\mu}_{\boldsymbol{\Psi}^{'}_{j}}, \sigma^2 = \hat{\sigma}^2_{\boldsymbol{\Psi}^{'}_{j}}),
	\end{aligned}	}
\end{equation*}where $\hat{\mu}_{\boldsymbol{\Psi}^{'}_{j}} = g_1(\mathbf{X})$ and $\hat{\sigma}^2_{\boldsymbol{\Psi}^{'}_{j}} = g_2(\mathbf{X})$.

The generative models for $\mathbf{X}$ and $Y$ vary depending on the types of variable. To be specific, for continuous $\mathbf{X}$ and $Y$, the models are defined as: 
\begin{equation*}
{\small 	\begin{aligned}
		p(\mathbf{X} | \boldsymbol{\Psi}) &= \prod_{j=1}^{D_{\mathbf{X}}}\mathcal{N}(\mu = \hat{\mu}_{\mathbf{X}_{j}}, \sigma^2 = \hat{\sigma}^2_{\mathbf{X}_{j}});\\
		\hat{\mu}_{\mathbf{X}_{j}} &= g_3(\boldsymbol{\Psi});~\hat{\sigma}^2_{\mathbf{X}_{j}} = g_4(\boldsymbol{\Psi}),
	\end{aligned}	}
\end{equation*}

\begin{equation*}
{\small 	\begin{aligned}
		p(Y | \boldsymbol{\Psi}) &= \prod_{j=1}^{D_{Y}}\mathcal{N}(\mu = \hat{\mu}_{Y_{j}}, \sigma^2 = \hat{\sigma}^2_{Y_{j}});\\
		\hat{\mu}_{Y_{j}} &= g_5(\boldsymbol{\Psi});~\hat{\sigma}^2_{Y_{j}} = g_6(\boldsymbol{\Psi}).
	\end{aligned}	}
\end{equation*}

For binary $\mathbf{X}$ and $Y$, the models are defined as: 
\begin{equation*}
	{\small\begin{aligned}
		p(\mathbf{X} | \boldsymbol{\Psi}) &= \mathrm{Bern}(\sigma(g_7(\boldsymbol{\Psi})));\\p(Y | \boldsymbol{\Psi})&= \mathrm{Bern}(\sigma(g_8(\boldsymbol{\Psi}))),
	\end{aligned}}
\end{equation*}where $\sigma(\cdot)$ is the logistic function.

We can present the evidence lower bound (ELBO) for the above inference and generative models:
\begin{equation*}
	\begin{aligned}
		\mathcal{M} =~ &\mathbb{E}_{q}[\log p(\mathbf{X} | \boldsymbol{\Psi})+ \log p(Y | \boldsymbol{\Psi})] \\&- D_{\mathrm{KL}}[q(\boldsymbol{\Psi} | \mathbf{X},Y)||p(\boldsymbol{\Psi} | \mathbf{X})],
		\label{eqa:ELBO}
	\end{aligned}
\end{equation*}where $D_{\mathrm{KL}}[\cdot||\cdot]$ is the $\mathrm{KL}$ divergence term. 

The first term denotes the reconstruction error of the observed $(\mathbf{X},Y)$ and the inferred $(\hat{\mathbf{X}},\hat{Y})$; other terms are used to calculate the $\mathrm{KL}$ divergence between the prior knowledge and the learned representations.

Following the works in~\cite{LouizosSMSZW17,ZhangLL21}, we design two auxiliary predictors that help us predict $T$ and $Y$ for new samples. To optimise the parameters in the auxiliary predictors, we add them to the loss function. Since maximising $\mathcal{M}$ is equal to minimising $-\mathcal{M}$, the final loss function of FDVAE is defined as:
\begin{equation*}
	\begin{aligned}
		\mathcal{L}_{\text{FDVAE}} =~ -\mathcal{M} &+ \mathbb{E}_{q}[\mathrm{log} q(T|\mathbf{X})] + \mathbb{E}_{q}[\mathrm{log} q(Y|\mathbf{X})].
	\end{aligned}
\end{equation*}

\section{Experiment}
Evaluating estimated causal effects is considered a great challenge, especially when there are unobserved confounders. Therefore, the evaluation of estimated causal effects with unobserved confounders relies on synthetic datasets since in this case we can obtain ground truth. Following the works in~\cite{LouizosSMSZW17,cheng2022toward}, we first generate some simulation datasets to compare the performance of FDVAE with other benchmark models in estimating causal effects, and then we demonstrate the strong correlation between the learned representations and the real FD adjustment sets. We validate that FDVAE can unbiasedly estimate the causal effects when changing the causal strength of unobserved confounders, and we also demonstrate the feasibility of FDVAE when the dimensionality of representation is mismatched with the dimensionality of the real FD adjustment set. Finally, we apply FDVAE to real datasets and verify its usability. The details of data generation and implementation of FDVAE are provided in the supplementary material.

\subsection{Experiment Setup}
\subsubsection{Models for Comparison}
We compare FDVAE with a number of benchmark models, including traditional and VAE based causal effect estimation models. The details of compared models are shown in Table~\ref{tab:models}, where the implementations of CEVAE and TEDVAE are retrieved from the authors’ GitHub and the implementations of other methods are from EconML~\cite{econml}.

\begin{table}[t]
	\centering
	{\small 	
	\begin{tabular}{ccc}
		\toprule
		Name 	& Reference	& Open-Source          \\ \midrule
		CEVAE	&\cite{LouizosSMSZW17}	& \href{https://github.com/AMLab-Amsterdam/CEVAE}{GitHub} \\
		LinearDRL	&\cite{chernozhukov2018double}	& \href{https://econml.azurewebsites.net/_autosummary/econml.dml.LinearDML.html}{EconML} \\
		CausalForest	&\cite{wager2018estimation}	& \href{https://econml.azurewebsites.net/_autosummary/econml.dml.CausalForestDML.html}{EconML} \\
		ForestDRL	&\cite{athey2019generalized}	& \href{https://econml.azurewebsites.net/_autosummary/econml.dr.ForestDRLearner.html}{EconML} \\
		XLearner	&\cite{kunzel2019metalearners}	& \href{https://econml.azurewebsites.net/_autosummary/econml.metalearners.XLearner.html}{EconML} \\
		KernelDML	&\cite{nie2021quasi}	& \href{https://econml.azurewebsites.net/_autosummary/econml.dml.KernelDML.html}{EconML} \\
		TEDVAE	&\cite{ZhangLL21}	& \href{https://github.com/WeijiaZhang24/TEDVAE}{GitHub} \\ \bottomrule
	\end{tabular}}
	\caption{Models for comparison.}
	\label{tab:models}
\end{table}

\begin{figure}[t]
	\centering
	\begin{subfigure}[l]{0.235\textwidth}
		\centering
		\includegraphics[scale=0.45]{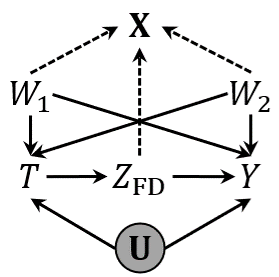}
		\caption{}
		\label{pic:settingA}
	\end{subfigure}
	\begin{subfigure}[r]{0.235\textwidth}
		\centering
		\includegraphics[scale=0.45]{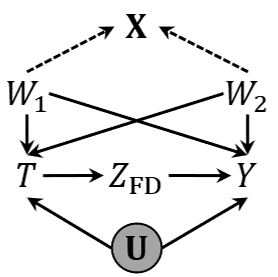}
		\caption{}
		\label{pic:settingB}
	\end{subfigure}
	\caption{(a) Setting A, where $\mathbf{X}$ is proxy of $\{W_1,W_2,Z_{\mathrm{FD}}\}$. \\(b) Setting B, where $\mathbf{X}$ is the proxy of $\{W_1,W_2\}$.}
	\label{pic:setting}
\end{figure}

\subsubsection{Evaluation Metrics}
For evaluating the performance of FDVAE and the other benchmark models, we use the Estimation Bias $|(\hat{\beta} - \beta)/\beta|\times100\%$ as the metric, where $\hat{\beta}$ is the estimated ATE and $\beta$ is the ground truth. 

\subsection{Experiments on Synthetic Datasets}
We generate synthetic datasets to evaluate the performance of FDVAE and the other comparison models. To avoid the bias brought by the data generation process, we repeatedly generate 30 datasets with a range of sample sizes (denoted as $N$), including 4k, 6k, 8k, 10k, 20k and 50k.

In order to better compare performance, we use different settings. Specifically, setting A is a common setting (as shown in Figure~\ref{pic:settingA}), where $\mathbf{X}$ is the proxy of $\{W_1,W_2,Z_{\mathrm{FD}}\}$. We can access $\mathbf{X}$ which contains the information of $Z_{\mathrm{FD}}$. In this setting, our proposed FDVAE method learns the representation of $Z_{\mathrm{FD}}$ from $\mathbf{X}$, and the comparison models use $\mathbf{X}$ as an adjustment set to estimate causal effects. Setting B (as shown in Figure~\ref{pic:settingB}) is a special setting that is designed for comparison models. This setting assumes $Z_{\mathrm{FD}}$ can be identified from $\mathbf{X}$. The comparison models use $\mathbf{X}$ as an adjustment set, where $\mathbf{X}$ is proxy of $\{W_1,W_2\}$. 

\begin{figure}[t]
	\centering
	\includegraphics[scale=0.655]{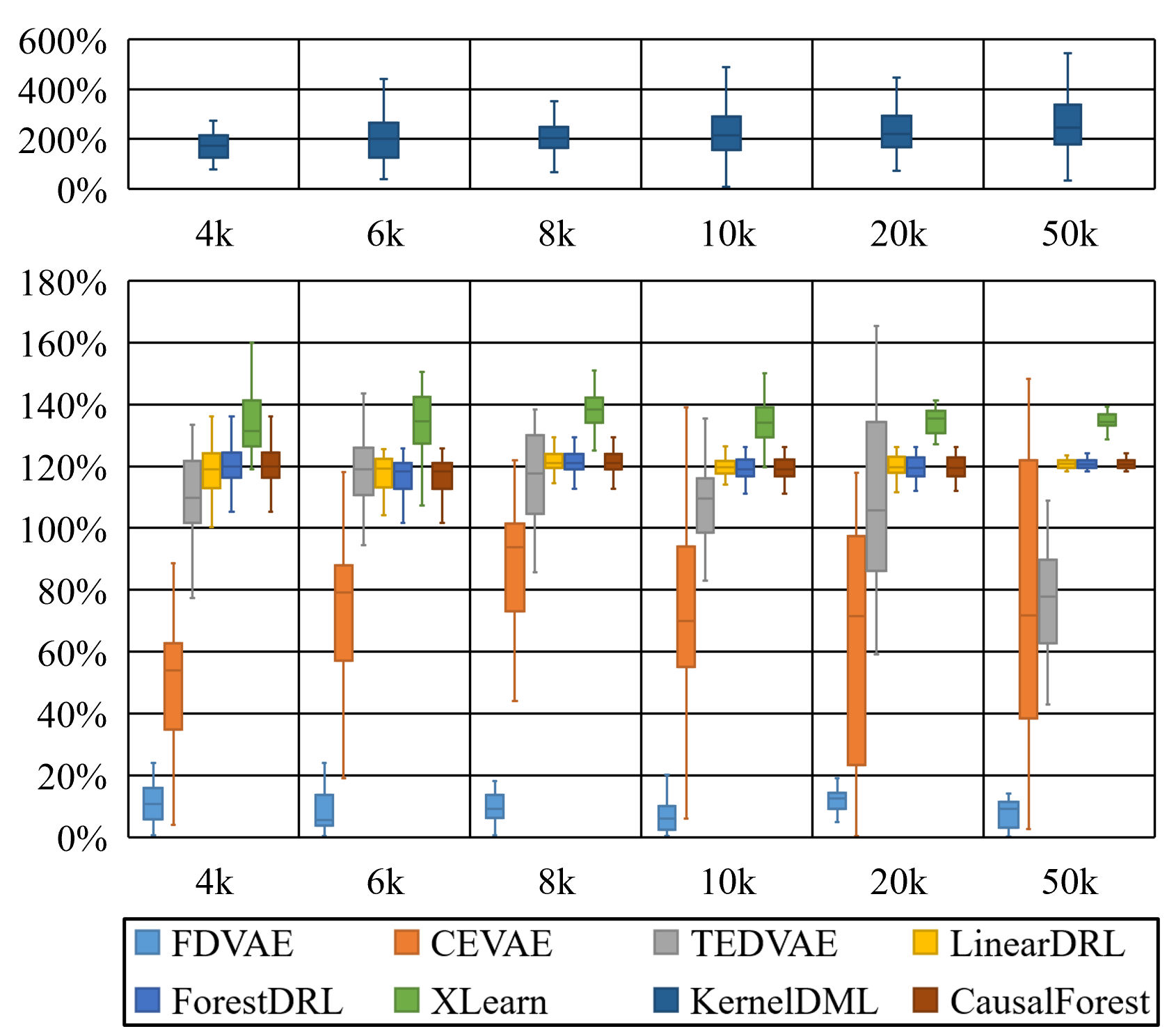}
	\caption{Results of Setting A, where the horizontal axis represents the sample size and the vertical axis represents the estimation bias (\%). The KernelDML model is displayed separately because it has a huge bias.}
	\label{pic:RES_settingA}
\end{figure}

\subsubsection{Results for Setting A}As shown in Figure~\ref{pic:RES_settingA}, FDVAE outperforms all the other comparison methods. Such results are expected. All comparison models use the BD criterion to estimate ATE. $Z_{\mathrm{FD}}$ produces a bias when being used as a part of the BD adjustment set, since $Z_{\mathrm{FD}}$ is a descendant node of $T$. On the other hand, the bias is also due to unobserved confounders $\mathbf{U}$. The unbiased causal effect estimation based on the BD criterion needs to block all the back-door paths from $T$ to $Y$. However, in our case, $\mathbf{U}$ cannot be used for adjustment because it is unobserved, and thus biased estimations are produced. 

FDVAE circumvents the limitations of the BD criterion. It learns the representation $\boldsymbol{\Psi}$ of the FD adjustment set from $\mathbf{X}$, and uses it to obtain the unbiased estimate of ATE.

\subsubsection{Results for Setting B} In setting B, since $\mathbf{X}$ is not a proxy of $Z_{\mathrm{FD}}$, using $\mathbf{X}$ for BD adjustment will not result in bias caused by $Z_{\mathrm{FD}}$ (which is a descendant of $T$ and is prohibited from being included in a BD adjustment set). Compared with the results of Setting A, all the comparison models have a significant performance improvement, and the biases of most of the models have reduced from nearly $120\%$ to about $30\%$. Under this setting, however, the estimation bias caused by unobserved confounders $\mathbf{U}$ cannot be eliminated. 

\begin{figure}[t]
	\centering
	\includegraphics[scale=0.655]{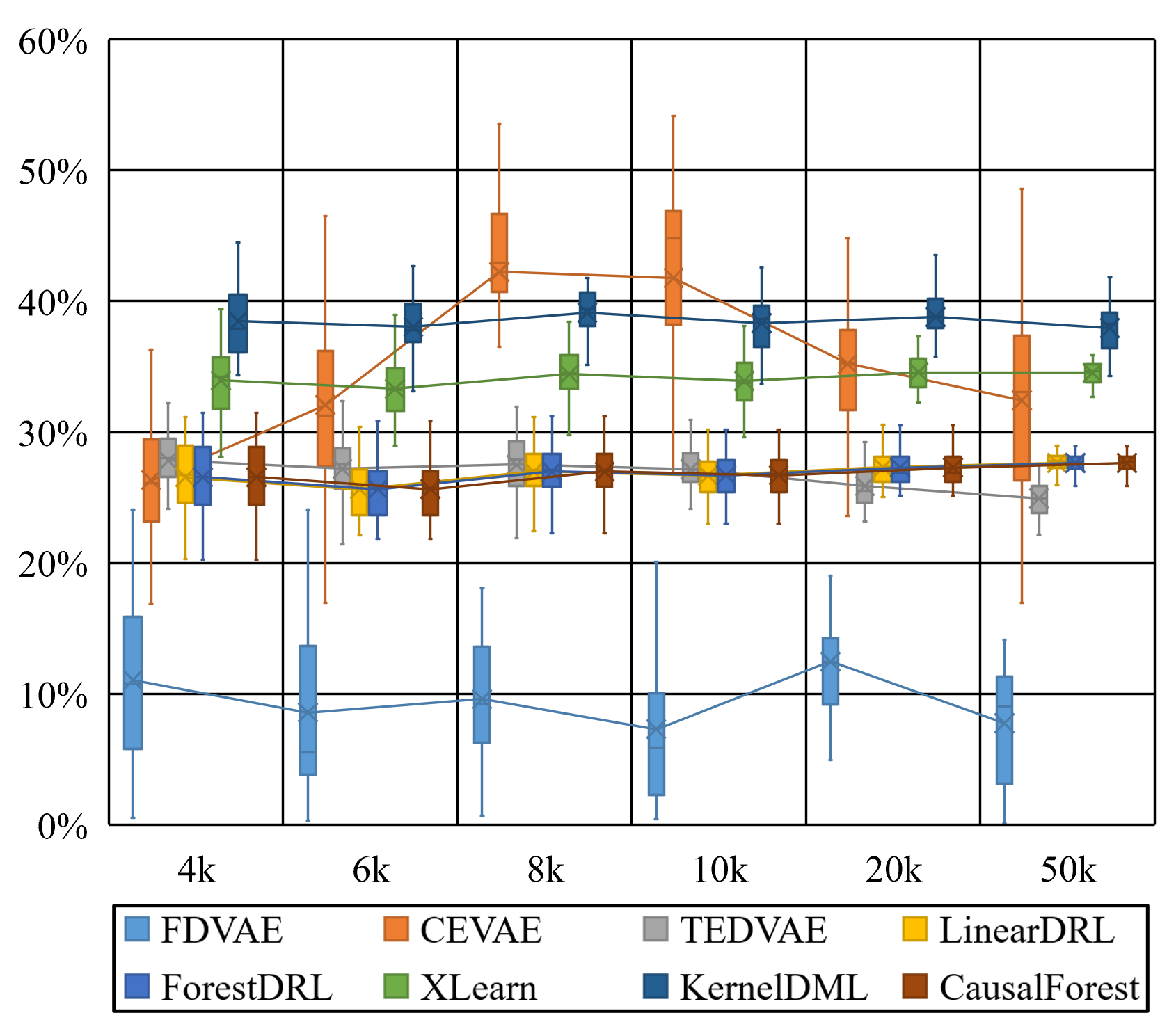}
	\caption{Results of Setting B. The line represents the mean of estimation bias, which is used to demonstrate the performance more clearly.}
	\label{pic:RES_settingB}
\end{figure}

As shown in Figure~\ref{pic:RES_settingB}, with the increase in sample size, the performance of our model is far better than the comparison models. This means that estimating the causal effect based on the FD criterion can avoid the bias caused by unobserved confounders.

\subsection{Effectiveness of the learned representation $\boldsymbol{\Psi}$} We note that the generation process of $\boldsymbol{\Psi}$ is regulated by given $\mathbf{X}$. In this section, we conduct experiments to demonstrate the effectiveness of this representation learning strategy. Since we use synthetic datasets, we know the ground truth value of $\mathbf{Z}_{\mathrm{FD}}$, and our proposed method FDVAE learns the representation $\boldsymbol{\Psi}$ of $\mathbf{Z}_{\mathrm{FD}}$. Both $D_{\mathbf{Z}_{\mathrm{FD}}}$ and $D_{\boldsymbol{\Psi}}$ are set to be $1$. After the representation learning process, we can get the learned representation $\boldsymbol{\psi}$ for each instance. 

In the empirical evaluation, we use the Pearson Correlation Coefficient (PCC) as the evaluation metric. We consider two variables are highly related when PCC is higher than $0.98$. As shown in Table~\ref{tab:PCC}, the obtained $\boldsymbol{\Psi}$ is highly related to $\mathbf{Z}_{\mathrm{FD}}$, which indicates that the representation learning strategy is effective in obtaining a proper FD adjustment set.

\begin{table}[t]
	\centering
	{\small\begin{tabular}{ccccc}
		\toprule
		N & PCC$(Z_{\mathrm{FD}},\Psi)$ & & N & PCC$(Z_{\mathrm{FD}},\Psi)$  \\ \cmidrule{1-2} \cmidrule{4-5}
		4k & 0.9986 ± 0.0009 & & 10k & 0.9982 ± 0.0017\\
		6k & 0.9979 ± 0.0029 & & 20k & 0.9944 ± 0.0164\\
		8k & 0.9984 ± 0.0018 & & 50k & 0.9897 ± 0.0096\\ \bottomrule
	\end{tabular}}
	\caption{The PCC (mean ± std) of the pair $(Z_{\mathrm{FD}},\Psi)$ for FDVAE under different sample size (denote as $N$).}
	\label{tab:PCC}
\end{table}

\subsection{Impact of the Causal Strength of $\mathbf{U}$}
\begin{figure}[t]
	\centering
	\includegraphics[scale=0.655]{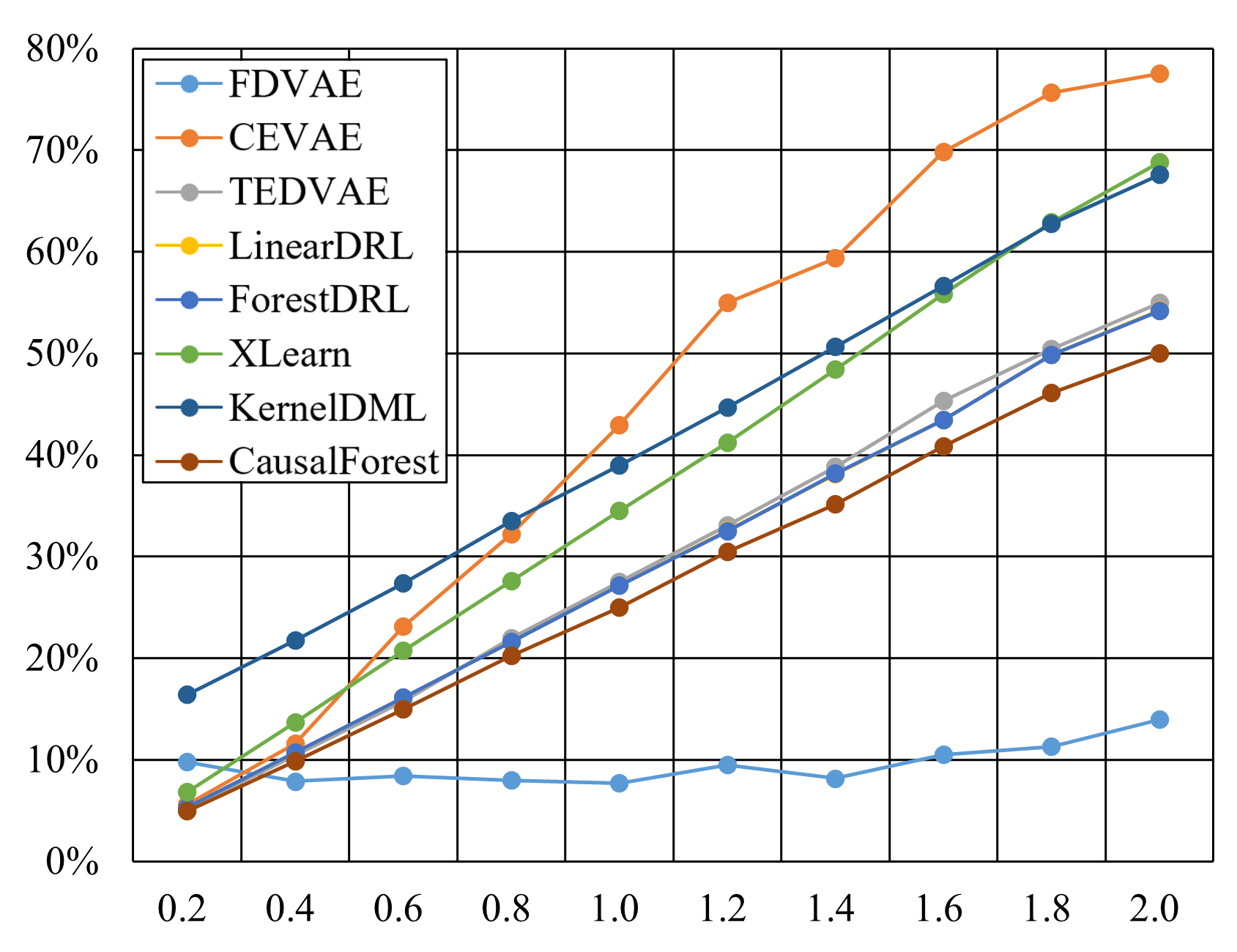}
	\caption{Results of different scale factor, where the horizontal axis represents the scale factor and the vertical axis represents the estimation bias (\%).}
	\label{pic:RES_U}
\end{figure}

We also conduct experiments to verify the effectiveness of FDVAE on the different causal strength of unobserved confounders. Since our data generation is based on SCMs, the causal strength of unobserved confounders can be varied by adjusting the coefficient of the path $\mathbf{U} \to Y$. The sample size for this experiment is fixed at 10k. We multiply a scale factor on the coefficient to realise the different causal strength levels of unobserved confounders. For example, $0.0$ means that there are no unobserved confounders, and $2.0$ means that the coefficient doubles the base value. We set the scaling range as $[0.0, 2.0]$ and the step increment as $0.2$. In this experiment, the comparison models are under Setting B, the setting is in favour of the comparison models, since under this setting when they use $\mathbf{X}$ as the back-door adjustment set does not result in estimation bias because $\mathbf{X}$ is not a proxy of $Z_{\mathrm{FD}}$. 

The results are shown in Figure~\ref{pic:RES_U}. As the scale factor increases, i.e., the causal strength of unobserved confounders increases, we notice that the estimation bias of the comparison models increases. FDVAE maintains an estimation bias of around $10\%$, with a slight upward trend, but much less pronounced compared to the comparison models. The results show that FDVAE is able to deal with unobserved confounders well, even when the causal strength of the unobserved confounders is high.

When the causal strength of the unobserved confounders is zero, that is, no unobserved confounders, the ATE can be estimated unbiasedly based on the BD criterion so the comparison models except KernelDML have minimal estimation bias. However, FDVAE still achieves acceptable performance, with a $10\%$ estimation bias due to the performance bottleneck of our representation learning caused by VAE.

\begin{table}[t]
	\centering
	{\small\begin{tabular}{ccccc}
			\toprule
			$D_{\mathbf{Z}_{\mathrm{FD}}}$ & Estimation Bias & &$D_{\mathbf{Z}_{\mathrm{FD}}}$ & Estimation Bias\\ \cmidrule{1-2} \cmidrule{4-5}
			1 & 7.55\% ± 4.71\% & & 6 & 11.43\% ± 7.06\% \\
			2 & 8.90\% ± 5.55\% & & 7 & 12.71\% ± 7.44\% \\
			3 & 7.75\% ± 5.76\% & & 8 & 12.39\% ± 6.99\%\\
			4 & 9.77\% ± 6.44\% & & 9 & 12.23\% ± 8.45\%\\
			5 & 9.65\% ± 6.76\% & & 10 & 11.75\% ± 8.51\%\\ \bottomrule
	\end{tabular}}
	\caption{The estimation bias (mean ± std) of FDVAE under the different settings of $D_{\mathbf{Z}_{\mathrm{FD}}}$.}
	\label{tab:DIM}
\end{table}

\subsection{Sensitivity to Representation Dimensionality} In real-world applications, it is a common situation that the dimensionality of representation is mismatched with the dimensionality of the real FD adjustment set. In this section, we use synthetic datasets with 10k samples and fix the dimensionality of representation as 1 (i.e., {\small $D_{\boldsymbol{\Psi}} = 1$}). We change the dimensionality of the real FD adjustment set from 1 to 10 (i.e., {\small $D_{\mathbf{Z}_{\mathrm{FD}}} = \{1,...,10\}$}) and evaluate the estimation of ATE. 

The results are shown in Table~\ref{tab:DIM}. We see that FDVAE achieves its best performance when the dimensionality of the representation matches the dimensionality of the original FD adjustment set (i.e., {\small $D_{\boldsymbol{\Psi}} = D_{\mathbf{Z}_{\mathrm{FD}}}$}). When the dimensionality of representation does not match the dimensionality of the real FD adjustment set, with the increase of $D_{\mathbf{Z}_{\mathrm{FD}}}$, the estimation bias is around $11\%$ and the std (standard deviation of the estimation bias) has an upward trend. However, compared to Figure~\ref{pic:RES_settingB}, even with dimensionality mismatch, the estimation bias of FDVAE is much lower than that of the compared models. In summary, FDVAE is not sensitive to the dimensionality of the representation and it can still perform relatively well when the dimensionality of the representation does not match that of the real FD adjustment set.

\begin{table}[t]
	\centering
	{\small\begin{tabular}{ccc}
			\toprule
			Dataset          & Empirical Intervals & FDVAE  \\ \midrule
			Sachs            & (0.05,3.23)             & 1.40   \\ 
			401k             & (0.047,0.095)           & 0.068  \\
			SchoolingReturns & (0.0484, 0.2175)        & 0.1199 \\ \bottomrule
	\end{tabular}}
	\caption{Results of real-world datasets.}
	\label{tab:case}
\end{table}

\subsection{Case Study on Real-World Datasets}
In this section, we apply FDVAE to the following three real-world datasets. As there are no ground truth causal effect values available for real-world datasets, for these datasets we use the commonly accepted empirical values in literature as the reference causal effect values.

\textbf{Sachs}: The dataset contains 853 samples and 11 variables~\cite{sachs2005causal}. The treatment is $Erk$, and the outcome is the concentration of $Akt$. The reference causal effect is $1.4301$ with $95\%$ confidence interval $(0.05,3.23)$~\cite{silva2017learning}.

\textbf{401k}: The dataset contains 9,275 individuals and 11 variables, and it is from the survey of income and program participation (SIPP)~\cite{chernozhukov2004impact}. The treatment is $p401k$, and the outcome is $pira$. The reference causal effect is $0.0712$ with $95\%$ confidence interval $(0.047,0.095)$~\cite{verbeek2008guide}.

\textbf{SchoolingReturns}: The dataset consists of 3,010 records and 19 variables~\cite{card1995using}. The treatment is the education level of a person, and the outcome is raw wages in 1976. The reference causal effect is $0.1329$ with $95\%$ confidence interval $(0.0484, 0.2175)$~\cite{verbeek2008guide}.

The results are shown in Table~\ref{tab:case}. We see that the estimated causal effects by FDVAE on the three real-world datasets are within their empirical intervals, implying the great potential of FDVAE for real-world applications.

\section{Related Work}
In this section, we review the research that is related to our work. Causal effect estimation has always been a challenging problem. Over the past few decades, researchers have proposed many methods for estimating causal effects. These methods generally fall into three categories: back-door adjustment, instrumental variable (IV) and front-door adjustment methods.

Methods based on the back-door criterion~\cite{pearl2009causality} are the most widely used, and most of these methods need to assume that all confounding variables are observed. For example, several tree-based models~\cite{athey2016recursive,su2009subgroup,zhang2017mining} have been designed to estimate causal effects by designing specific splitting criterion; meta-learning~\cite{kunzel2019metalearners} has also been proposed to utilise existing machine learning algorithms for causal effect estimation. Recently, methods using deep learning techniques to predict causal effects have received widespread attention. For example, CEVAE~\cite{LouizosSMSZW17} combines representation learning and VAE to estimate causal effects; TEDVAE~\cite{ZhangLL21} improves on CEVAE and decouples the learned representations to achieve more accurate estimation; Counterfactual regression nets~\cite{johansson2016learning,shalit2017estimating,hassanpour2019counterfactual} proposed to balance treated and untreated sample groups so that the two groups are as close as possible.

Methods based on instrumental variables (IV) have also received a lot of attention. IV based methods are able to solve the problem of unobserved confounding variables. Most IV based methods require users to nominate a valid IV, such as the generalised method of moments (GMM)~\cite{bennett2019deep}, kernel-IV regression~\cite{singh2019kernel} and deep learning based method~\cite{hartford2017deep}. When there are no nominated IVs in the data, some data-driven methods are developed to find~\cite{yuan2022auto} or synthesise~\cite{burgess2013use,kuang2020ivy} an IV or eliminate the influence of invalid IVs by using statistical strategies~\cite{guo2018confidence,hartford2021valid}. 

However, the front-door criterion based approach is rarely studied in the literature. There are only a few methods for finding the appropriate adjustment set by following the front-door criterion in the observed variables~\cite{jeong2022finding,wienobst2022finding}. These methods aim to find and enumerate possible sets that satisfying the front-door criterion in a given causal graph. Therefore, these methods are not efficient when dealing with high-dimensional data.

In summary, the methods based on the back-door criterion cannot handle unobserved confounders. IV based methods can cope with unobserved confounders, but cannot deal with the situation where the treatment affects the outcome through a mediator. Our method can simultaneously address the existence of unobserved confounders and mediators, and it is the first data-driven causal effect estimation method based on the FD criterion.

\section{Conclusion}
This work studies the problem of causal effect estimation from observational data when there exist unobserved confounders and the treatment does not directly affect the outcome but through a mediator. We have proposed the first data-driven method, FDVAE, for dealing with the problem. We use VAE technique to learn the representation of the front-door adjustment set from proxy variables. Extensive experiments have demonstrated that our proposed method outperforms other methods in the presence of unobserved confounders and mediators. We have also shown that our method is insensitive to the causal strength of unobserved confounders in the data. The performance of other methods is significantly worse as the causal strength of the unobserved confounders increases. Furthermore, we have validated the usability of FDVAE on three real-world datasets.

%\section*{Ethical Statement}
%
%There are no ethical issues.
%
%\section*{Acknowledgments}
%
%The preparation of these instructions and the \LaTeX{} and Bib\TeX{}
%files that implement them was supported by Schlumberger Palo Alto
%Research, AT\&T Bell Laboratories, and Morgan Kaufmann Publishers.
%Preparation of the Microsoft Word file was supported by IJCAI.  An
%early version of this document was created by Shirley Jowell and Peter
%F. Patel-Schneider.  It was subsequently modified by Jennifer
%Ballentine and Thomas Dean, Bernhard Nebel, Daniel Pagenstecher,
%Kurt Steinkraus, Toby Walsh and Carles Sierra. The current version
%has been prepared by Marc Pujol-Gonzalez and Francisco Cruz-Mencia.

\newpage
\bibliographystyle{named}
\bibliography{FDVAE}

\begin{thebibliography}{}

\bibitem[\protect\citeauthoryear{Ascarza}{2018}]{ascarza2018retention}
Eva Ascarza.
\newblock Retention futility: Targeting high-risk customers might be
  ineffective.
\newblock {\em Journal of Marketing Research}, 55(1):80--98, 2018.

\bibitem[\protect\citeauthoryear{Athey and Imbens}{2016}]{athey2016recursive}
Susan Athey and Guido Imbens.
\newblock Recursive partitioning for heterogeneous causal effects.
\newblock {\em Proceedings of the National Academy of Sciences},
  113(27):7353--7360, 2016.

\bibitem[\protect\citeauthoryear{Athey \bgroup \em et al.\egroup
  }{2019}]{athey2019generalized}
Susan Athey, Julie Tibshirani, and Stefan Wager.
\newblock Generalized random forests.
\newblock {\em The Annals of Statistics}, 47(2):1148--1178, 2019.

\bibitem[\protect\citeauthoryear{Athey}{2017}]{athey2017beyond}
Susan Athey.
\newblock Beyond prediction: Using big data for policy problems.
\newblock {\em Science}, 355(6324):483--485, 2017.

\bibitem[\protect\citeauthoryear{Bareinboim and
  Pearl}{2016}]{bareinboim2016causal}
Elias Bareinboim and Judea Pearl.
\newblock Causal inference and the data-fusion problem.
\newblock {\em Proceedings of the National Academy of Sciences},
  113(27):7345--7352, 2016.

\bibitem[\protect\citeauthoryear{Bennett \bgroup \em et al.\egroup
  }{2019}]{bennett2019deep}
Andrew Bennett, Nathan Kallus, and Tobias Schnabel.
\newblock Deep generalized method of moments for instrumental variable
  analysis.
\newblock In {\em Advances in Neural Information Processing Systems 32,
  {NIPS}}, pages 3559--3569, 2019.

\bibitem[\protect\citeauthoryear{Burgess and Thompson}{2013}]{burgess2013use}
Stephen Burgess and Simon~G Thompson.
\newblock Use of allele scores as instrumental variables for mendelian
  randomization.
\newblock {\em International Journal of Epidemiology}, 42(4):1134--1144, 2013.

\bibitem[\protect\citeauthoryear{Card}{1995}]{card1995using}
David Card.
\newblock Using geographical variation in college proximity to estimate the
  re\& turn to schoolingv.
\newblock {\em Aspects of Labor Market Behavior: Essays in Honor of John
  Vanderkamp, University of Toronto Press, Toronto}, 1995.

\bibitem[\protect\citeauthoryear{Cheng \bgroup \em et al.\egroup
  }{2022a}]{cheng2022toward}
Debo Cheng, Jiuyong Li, Lin Liu, Kui Yu, Thuc~Duy Le, and Jixue Liu.
\newblock Toward unique and unbiased causal effect estimation from data with
  hidden variables.
\newblock {\em IEEE Transactions on Neural Networks and Learning Systems},
  pages 1--13, 2022.

\bibitem[\protect\citeauthoryear{Cheng \bgroup \em et al.\egroup
  }{2022b}]{cheng2022causal}
Lu~Cheng, Ruocheng Guo, and Huan Liu.
\newblock Causal mediation analysis with hidden confounders.
\newblock In {\em The Fifteenth {ACM} International Conference on Web Search
  and Data Mining, {WSDM}}, pages 113--122, 2022.

\bibitem[\protect\citeauthoryear{Chernozhukov and
  Hansen}{2004}]{chernozhukov2004impact}
Victor Chernozhukov and Christian Hansen.
\newblock The impact of 401 (k) participation on the wealth distribution: An
  instrumental quantile regression analysis.
\newblock {\em Review of Economics and statistics}, 86(3):735--751, 2004.

\bibitem[\protect\citeauthoryear{Chernozhukov \bgroup \em et al.\egroup
  }{2018}]{chernozhukov2018double}
Victor Chernozhukov, Denis Chetverikov, Mert Demirer, Esther Duflo, Christian
  Hansen, Whitney Newey, and James Robins.
\newblock Double/debiased machine learning for treatment and structural
  parameters: Double/debiased machine learning.
\newblock {\em The Econometrics Journal}, 21(1), 2018.

\bibitem[\protect\citeauthoryear{Deaton and
  Cartwright}{2018}]{deaton2018understanding}
Angus Deaton and Nancy Cartwright.
\newblock Understanding and misunderstanding randomized controlled trials.
\newblock {\em Social Science \& Medicine}, 210:2--21, 2018.

\bibitem[\protect\citeauthoryear{Fisher}{1936}]{fisher1936design}
Ronald~Aylmer Fisher.
\newblock Design of experiments.
\newblock {\em British Medical Journal}, 1(3923):554, 1936.

\bibitem[\protect\citeauthoryear{Guo \bgroup \em et al.\egroup
  }{2018}]{guo2018confidence}
Zijian Guo, Hyunseung Kang, T~Tony~Cai, and Dylan~S Small.
\newblock Confidence intervals for causal effects with invalid instruments by
  using two-stage hard thresholding with voting.
\newblock {\em Journal of the Royal Statistical Society: Series B (Statistical
  Methodology)}, 80(4):793--815, 2018.

\bibitem[\protect\citeauthoryear{Hartford \bgroup \em et al.\egroup
  }{2017}]{hartford2017deep}
Jason Hartford, Greg Lewis, Kevin Leyton-Brown, and Matt Taddy.
\newblock Deep {IV:} {A} flexible approach for counterfactual prediction.
\newblock In {\em Proceedings of the 34th International Conference on Machine
  Learning, {ICML}}, pages 1414--1423, 2017.

\bibitem[\protect\citeauthoryear{Hartford \bgroup \em et al.\egroup
  }{2021}]{hartford2021valid}
Jason~S Hartford, Victor Veitch, Dhanya Sridhar, and Kevin Leyton-Brown.
\newblock Valid causal inference with (some) invalid instruments.
\newblock In {\em Proceedings of the 38th International Conference on Machine
  Learning, {ICML}}, pages 4096--4106, 2021.

\bibitem[\protect\citeauthoryear{Hassanpour and
  Greiner}{2019}]{hassanpour2019counterfactual}
Negar Hassanpour and Russell Greiner.
\newblock Counterfactual regression with importance sampling weights.
\newblock In {\em Proceedings of the Twenty-Eighth International Joint
  Conference on Artificial Intelligence, {IJCAI}}, pages 5880--5887, 2019.

\bibitem[\protect\citeauthoryear{Hern{\'a}n and
  Robins}{2006}]{hernan2006instruments}
Miguel~A Hern{\'a}n and James~M Robins.
\newblock Instruments for causal inference: An epidemiologist's dream?
\newblock {\em Epidemiology}, pages 360--372, 2006.

\bibitem[\protect\citeauthoryear{Imbens and Rubin}{2015}]{imbens2015causal}
Guido~W Imbens and Donald~B Rubin.
\newblock {\em Causal Inference in Statistics, Social, and Biomedical
  Sciences}.
\newblock Cambridge University Press, 2015.

\bibitem[\protect\citeauthoryear{Jeong \bgroup \em et al.\egroup
  }{2022}]{jeong2022finding}
Hyunchai Jeong, Jin Tian, and Elias Bareinboim.
\newblock Finding and listing front-door adjustment sets.
\newblock {\em arXiv preprint arXiv:2210.05816}, 2022.

\bibitem[\protect\citeauthoryear{Johansson \bgroup \em et al.\egroup
  }{2016}]{johansson2016learning}
Fredrik Johansson, Uri Shalit, and David Sontag.
\newblock Learning representations for counterfactual inference.
\newblock In {\em Proceedings of the 33nd International Conference on Machine
  Learning, {ICML}}, pages 3020--3029, 2016.

\bibitem[\protect\citeauthoryear{Keith~Battocchi}{2019}]{econml}
Maggie Hei Greg Lewis Paul Oka Miruna Oprescu Vasilis~Syrgkanis
  Keith~Battocchi, Eleanor~Dillon.
\newblock {EconML}: {A Python Package for ML-Based Heterogeneous Treatment
  Effects Estimation}.
\newblock https://github.com/microsoft/EconML, 2019.
\newblock Version 0.13.

\bibitem[\protect\citeauthoryear{Kingma and Welling}{2014}]{KingmaW13}
Diederik~P. Kingma and Max Welling.
\newblock Auto-encoding variational bayes.
\newblock In {\em 2nd International Conference on Learning Representations,
  {ICLR}}, pages 1--14, 2014.

\bibitem[\protect\citeauthoryear{Kuang \bgroup \em et al.\egroup
  }{2020}]{kuang2020ivy}
Zhaobin Kuang, Frederic Sala, Nimit Sohoni, Sen Wu, Aldo C{\'o}rdova-Palomera,
  Jared Dunnmon, James Priest, and Christopher R{\'e}.
\newblock Ivy: Instrumental variable synthesis for causal inference.
\newblock In {\em The 23rd International Conference on Artificial Intelligence
  and Statistics, {AISTATS}}, pages 398--410, 2020.

\bibitem[\protect\citeauthoryear{K{\"u}nzel \bgroup \em et al.\egroup
  }{2019}]{kunzel2019metalearners}
S{\"o}ren~R K{\"u}nzel, Jasjeet~S Sekhon, Peter~J Bickel, and Bin Yu.
\newblock Metalearners for estimating heterogeneous treatment effects using
  machine learning.
\newblock {\em Proceedings of the National Academy of Sciences},
  116(10):4156--4165, 2019.

\bibitem[\protect\citeauthoryear{Louizos \bgroup \em et al.\egroup
  }{2017}]{LouizosSMSZW17}
Christos Louizos, Uri Shalit, Joris~M. Mooij, David~A. Sontag, Richard~S.
  Zemel, and Max Welling.
\newblock Causal effect inference with deep latent-variable models.
\newblock In {\em Advances in Neural Information Processing Systems 30,
  {NIPS}}, pages 6446--6456, 2017.

\bibitem[\protect\citeauthoryear{Nie and Wager}{2021}]{nie2021quasi}
Xinkun Nie and Stefan Wager.
\newblock Quasi-oracle estimation of heterogeneous treatment effects.
\newblock {\em Biometrika}, 108(2):299--319, 2021.

\bibitem[\protect\citeauthoryear{Pearl}{2009}]{pearl2009causality}
Judea Pearl.
\newblock {\em Causality}.
\newblock Cambridge university press, 2009.

\bibitem[\protect\citeauthoryear{Peters \bgroup \em et al.\egroup
  }{2017}]{peters2017elements}
Jonas Peters, Dominik Janzing, and Bernhard Sch{\"o}lkopf.
\newblock {\em Elements of Causal Inference: Foundations and Learning
  Algorithms}.
\newblock The MIT Press, 2017.

\bibitem[\protect\citeauthoryear{Petersen and van~der
  Laan}{2014}]{petersen2014causal}
Maya~L Petersen and Mark~J van~der Laan.
\newblock Causal models and learning from data: Integrating causal modeling and
  statistical estimation.
\newblock {\em Epidemiology (Cambridge, Mass.)}, 25(3):418, 2014.

\bibitem[\protect\citeauthoryear{Sachs \bgroup \em et al.\egroup
  }{2005}]{sachs2005causal}
Karen Sachs, Omar Perez, Dana Pe'er, Douglas~A Lauffenburger, and Garry~P
  Nolan.
\newblock Causal protein-signaling networks derived from multiparameter
  single-cell data.
\newblock {\em Science}, 308(5721):523--529, 2005.

\bibitem[\protect\citeauthoryear{Shalit \bgroup \em et al.\egroup
  }{2017}]{shalit2017estimating}
Uri Shalit, Fredrik~D Johansson, and David Sontag.
\newblock Estimating individual treatment effect: Generalization bounds and
  algorithms.
\newblock In {\em Proceedings of the 33nd International Conference on Machine
  Learning, {ICML}}, pages 3076--3085, 2017.

\bibitem[\protect\citeauthoryear{Silva and Shimizu}{2017}]{silva2017learning}
Ricardo Silva and Shohei Shimizu.
\newblock Learning instrumental variables with structural and non-gaussianity
  assumptions.
\newblock {\em Journal of Machine Learning Research}, 18(120):1--49, 2017.

\bibitem[\protect\citeauthoryear{Singh \bgroup \em et al.\egroup
  }{2019}]{singh2019kernel}
Rahul Singh, Maneesh Sahani, and Arthur Gretton.
\newblock Kernel instrumental variable regression.
\newblock In {\em Advances in Neural Information Processing Systems 32,
  {NIPS}}, pages 4595--4607, 2019.

\bibitem[\protect\citeauthoryear{Spirtes \bgroup \em et al.\egroup
  }{2000}]{spirtes2000causation}
Peter Spirtes, Clark~N Glymour, Richard Scheines, and David Heckerman.
\newblock {\em Causation, Prediction, and Search}.
\newblock MIT press, 2000.

\bibitem[\protect\citeauthoryear{Su \bgroup \em et al.\egroup
  }{2009}]{su2009subgroup}
Xiaogang Su, Chih-Ling Tsai, Hansheng Wang, David~M Nickerson, and Bogong Li.
\newblock Subgroup analysis via recursive partitioning.
\newblock {\em Journal of Machine Learning Research}, 10(2), 2009.

\bibitem[\protect\citeauthoryear{Tran \bgroup \em et al.\egroup
  }{2022}]{tran2022most}
Ha~Xuan Tran, Thuc~Duy Le, Jiuyong Li, Lin Liu, Jixue Liu, Yanchang Zhao, and
  Tony Waters.
\newblock What is the most effective intervention to increase job retention for
  this disabled worker?
\newblock In {\em Proceedings of the 28th ACM SIGKDD Conference on Knowledge
  Discovery and Data Mining, {KDD}}, pages 3981--3991, 2022.

\bibitem[\protect\citeauthoryear{Verbeek}{2008}]{verbeek2008guide}
Marno Verbeek.
\newblock {\em A Guide to Modern Econometrics}.
\newblock John Wiley \& Sons, 2008.

\bibitem[\protect\citeauthoryear{Wager and Athey}{2018}]{wager2018estimation}
Stefan Wager and Susan Athey.
\newblock Estimation and inference of heterogeneous treatment effects using
  random forests.
\newblock {\em Journal of the American Statistical Association},
  113(523):1228--1242, 2018.

\bibitem[\protect\citeauthoryear{Wien{\"o}bst \bgroup \em et al.\egroup
  }{2022}]{wienobst2022finding}
Marcel Wien{\"o}bst, Benito van~der Zander, and Maciej Li{\'s}kiewicz.
\newblock Finding front-door adjustment sets in linear time.
\newblock {\em arXiv preprint arXiv:2211.16468}, 2022.

\bibitem[\protect\citeauthoryear{Yuan \bgroup \em et al.\egroup
  }{2022}]{yuan2022auto}
Junkun Yuan, Anpeng Wu, Kun Kuang, Bo~Li, Runze Wu, Fei Wu, and Lanfen Lin.
\newblock Auto iv: Counterfactual prediction via automatic instrumental
  variable decomposition.
\newblock {\em ACM Transactions on Knowledge Discovery from Data (TKDD)},
  16(4):1--20, 2022.

\bibitem[\protect\citeauthoryear{Zhang \bgroup \em et al.\egroup
  }{2017}]{zhang2017mining}
Weijia Zhang, Thuc~Duy Le, Lin Liu, Zhi-Hua Zhou, and Jiuyong Li.
\newblock Mining heterogeneous causal effects for personalized cancer
  treatment.
\newblock {\em Bioinformatics}, 33(15):2372--2378, 2017.

\bibitem[\protect\citeauthoryear{Zhang \bgroup \em et al.\egroup
  }{2021}]{ZhangLL21}
Weijia Zhang, Lin Liu, and Jiuyong Li.
\newblock Treatment effect estimation with disentangled latent factors.
\newblock In {\em Thirty-Fifth {AAAI} Conference on Artificial Intelligence,
  {AAAI}}, pages 10923--10930, 2021.

\end{thebibliography}

\end{document}